\documentclass[10pt,twocolumn,letterpaper]{article}

\usepackage[pagenumbers]{cvpr} 

\usepackage{graphicx}
\usepackage{amsmath, amssymb} 
\usepackage{bm}
\usepackage{textcomp}
\usepackage{gensymb}
\usepackage{pifont}
\usepackage[T1]{fontenc}
\usepackage{multirow}
\usepackage{makecell}
\usepackage{xspace}
\usepackage{enumitem}
\usepackage{url}
\usepackage{enumerate}
\usepackage{algorithm}
\usepackage[noend]{algpseudocode} 
\usepackage{wrapfig,booktabs}
\makeatletter
\DeclareMathOperator*{\argmax}{arg\,max}
\DeclareMathOperator*{\argmin}{arg\,min}
\DeclareRobustCommand\onedot{\futurelet\@let@token\@onedot}
\def\@onedot{\ifx\@let@token.\else.\null\fi\xspace}
\def\eg{\emph{e.g}\onedot} 
\def\ie{\emph{i.e}\onedot} 
 
 \def\vs{\emph{vs}\onedot}

\def\ve{\boldsymbol}
\DeclareMathOperator{\Tr}{Tr}
\makeatother

\usepackage{hyperref}
\hypersetup{breaklinks,colorlinks}

\DeclareUnicodeCharacter{2212}{-}

\def\cvprPaperID{387} 
\def\confName{CVPR}
\def\confYear{2023}

\begin{document}
\title{Large-scale Training Data Search for Object Re-identification}


\author{Yue Yao$^1$  \hspace{0.8cm}  
Huan Lei$^1$  \hspace{0.8cm}  
Tom Gedeon$^{2}$  \hspace{0.8cm}  Liang Zheng$^1$ \\$^1$Australian National University  \hspace{0.4cm}   $^2$Curtin University \\ \{first name.second name\}@\{anu | curtin\}.edu.au
}

\maketitle

\begin{abstract}
We consider a scenario where we have access to the target domain, but cannot afford on-the-fly training data annotation, and instead would like to construct an alternative training set from a large-scale data pool such that a competitive model can be obtained. We propose a search and pruning (SnP) solution to this training data search problem, tailored to object re-identification (re-ID), an application aiming to match the same object captured by different cameras. Specifically, the search stage identifies and merges clusters of source identities which exhibit similar distributions with the target domain. The second stage, subject to a budget, then selects identities and their images from the Stage I output, to control the size of the resulting training set for efficient training. The two steps provide us with training sets 80\% smaller than the source pool while achieving a similar or even higher re-ID accuracy. These training sets are also shown to be superior to a few existing search methods such as random sampling and greedy sampling under the same budget on training data size. If we release the budget, training sets resulting from the first stage alone allow even higher re-ID accuracy. We provide interesting discussions on the specificity of our method to the re-ID problem and particularly its role in bridging the re-ID domain gap. The code is available at \url{https://github.com/yorkeyao/SnP}
    
\end{abstract}

\section{Introduction}
\label{sec:intro}

\begin{figure}[t]
\begin{center}
	\includegraphics[width=1\linewidth]{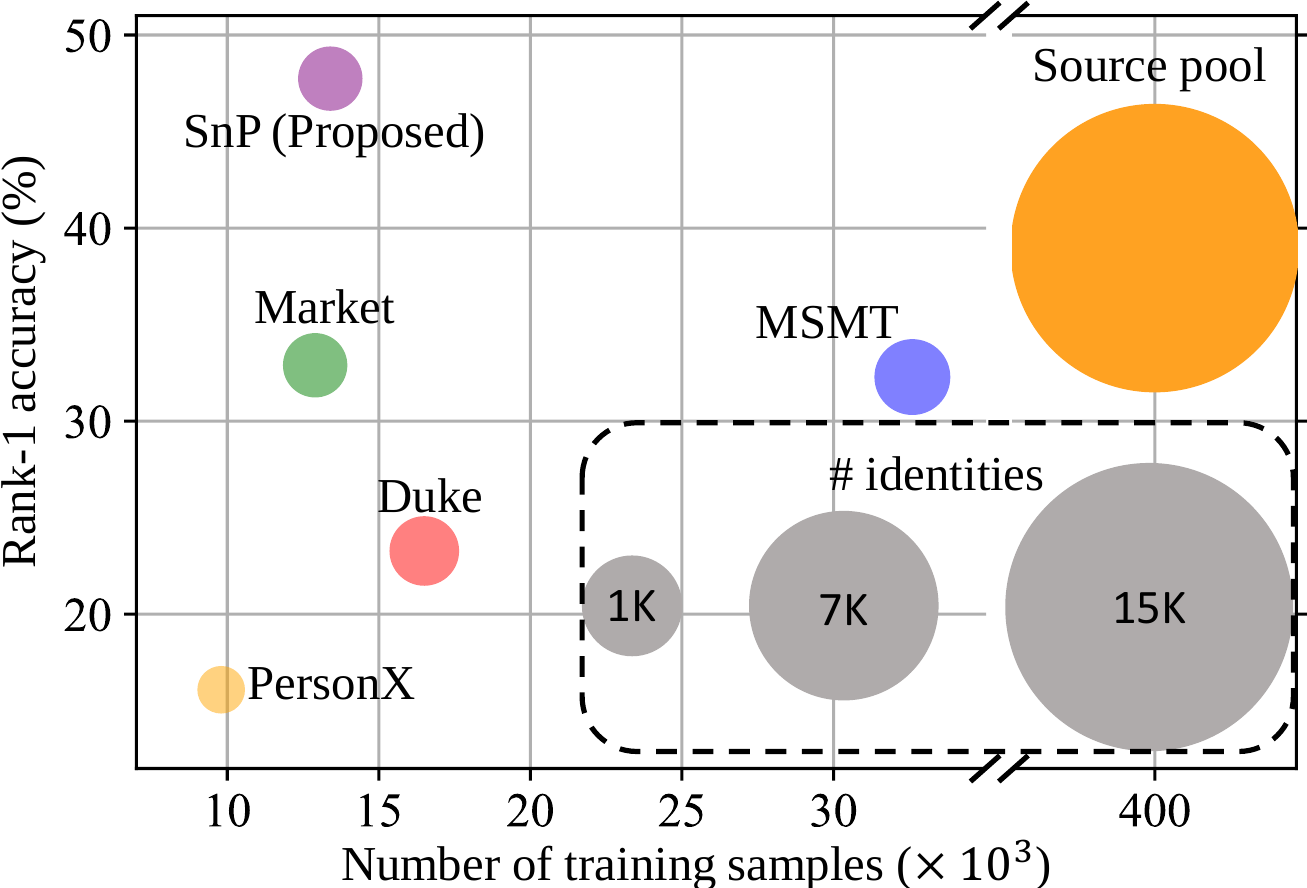}
\end{center}
\caption{We present a search and pruning (SnP) solution to the training data search problem in object re-ID. The source data pool is 1 order of magnitude larger than existing re-ID training sets in terms of the number of images and the number of identities. When the target is AlicePerson~\cite{alice_benchmark}, from the source pool, our method (SnP) results in a training set 80\% smaller than the source pool while achieving a similar or even higher re-ID accuracy. The searched training set is also superior to existing individual training sets such as Market-1501~\cite{zheng2015scalable}, Duke~\cite{zheng2017unlabeled}, and MSMT~\cite{wei2018person}. }
\label{fig:intro}
\end{figure}

The success of a deep learning-based object re-ID relies on one of its critical prerequisites: the labeled training data. To achieve high accuracy, typically a massive amount of data needs to be used to train deep learning models. However, creating large-scale object re-ID training data with manual labels is expensive. Furthermore, collecting training data that contributes to the high test accuracy of the trained model is even more challenging. Recent years have seen a large amount of datasets proposed and a significant increase in the data size of a single dataset. For example, the RandPerson~\cite{sun2019dissecting} dataset has 8000 identities, which is more than $6\times$ larger than the previous PersonX~\cite{sun2019dissecting} dataset.

However, these datasets generally have their own dataset bias, making the model trained on one dataset unable to generalize well to another. For example, depending on the filming scenario, different person re-ID datasets generally have biases on camera positions, human races, and clothing style. Such dataset biases usually lead to model bias, which results in the model's difficulty performing well in an unseen filming scenario. To address this, many try to improve learning algorithms, including domain adaptation and domain generalization methods~\cite{fan2018unsupervised,zhong2019invariance,song2020learning,song2019generalizable,luo2020generalizing,bai2021person30k}. Whereas these algorithms are well-studied and have proven successful in many re-ID applications, deciding what kind of data to use for training the re-ID model is still an open research problem, and has received relatively little attention in the community. We argue that this is a crucial problem to be answered in light of the ever-increasing scale of the available datasets.

In this paper, we introduce SnP, a search and pruning solution for sampling an efficient re-ID training set to a target domain. SnP is designed for the scenario in that we have a target dataset that does not have labeled training data. Our collected source pool, in replace, provides  suitable labeled data to train a competitive model. Specifically, given a user-specified budget (\eg, maximum desired data size), we sample a subset of the source pool, which satisfies the budget requirement and desirably has high-quality data to train deep learning models. This scenario is especially helpful for deploying a re-ID system for unknown test environments, as it is difficult to manually label a training set for these new environments. We note that due to the absence of an in-distribution training set, the searched data are directly used for training the re-ID model rather than pre-training.   

In particular, we combine several popular re-ID datasets into a source pool, and represent each image in the pool with features. Those features are extracted from an Imagenet-pretrained model~\cite{szegedy2016rethinking}. The images with features are stored on the dataserver to serve as a gallery. When there is a query from the target, we extract the feature of the query image, and search in the gallery for similar images on a feature level. Specifically, in the search stage, we calculate feature-level distance, \ie, Fr\'{e}chet Inception Distance (FID)~\cite{heusel2017gans}.  Given the constraint of a budget, we select the most representative samples in the pruning stage, based on the outputs from the search stage. This limits the size of the constructed training set  and enables efficient training. 

Combining search and pruning, we construct a training dataset that empirically shows significant accuracy improvements on several object re-ID targets, compared to the baselines. Without budget constraints, our searched training sets allow higher re-ID accuracy than the complete source pool, due to its target-specificity. With budget constraints, the pruned training set still achieves comparable or better performance than the source pool. The proposed SnP is demonstrated to be superior to random or greedy sampling. We show in Fig.~\ref{fig:intro} that the training set constructed by SnP leads to the best performance on the target compared to the others. We provide discussions on the specificity of our method and its role in bridging the re-ID domain gap.

\section{Related Work}
\textbf{Active learning} gradually searches over unlabeled data to find samples to be labeled by the oracle~\cite{settles2009active}. It is an iterative training process, where data search is performed at each iteration to train a task model. Our task is different from active learning. First, active learning is designed to acquire a training set and train models gradually. This requires multiple real training processes which are computationally expensive. In comparison, we directly search the whole training set all at once. Then the searched training set is used for model training. Second, many active learning methods require access to the target task model to provide selection metrics, \eg, uncertainty-based metrics~\cite{gal2017deep,yang2015multi,elhamifar2013convex,guo2010active}. Our task does not require task model information, in other words, a real training process, during the process of the training set search. Thus enabling a fast training data retrial.

\textbf{Neural data server} is the closest inspiring work~\cite{yan2020neural, settles2009active}. They also aim to search training data all at once from a large database. However, compared with us, firstly,~\cite{yan2020neural} and~\cite{settles2009active} are designed for searching pretraining data rather than direct training data. Such a design is understandable as they are mainly for the classification task. Searching direct training data require careful class alignment. In comparison, We are targeting the re-ID task, where its training set can have a different class from the target, and then be directly used for training models. Furthermore,~\cite{yan2020neural} and~\cite{settles2009active} require unsupervised pretrained experts to measure the domain gap. While we do not require so, which saves extraction time and simplify the solution.  


\textbf{Transfer learning} is a long-standing problem for re-ID tasks, and many attempts on learning algorithms have been made to reduce the effect of domain gap~\cite{torralba2011unbiased,perronnin2010improving,saenko2010adapting,deng2018image,lou2019embedding}. Common strategies contain feature-level~\cite{long2015learning} and pseudo-label based~\cite{fan2018unsupervised,zhong2019invariance,song2020learning} domain adaptation, and domain generalization~\cite{song2019generalizable,luo2020generalizing,bai2021person30k}. In this paper, we focus on training data that is orthogonal to existing training algorithms. To be shown in experiments, together with some domain adaptation (\ie, pseudo-label) methods, SnP can achieve higher re-ID accuracy. 

\textbf{Learning to generate synthetic training data.} Data simulation is also an inexpensive alternative to increasing a training set scale while providing accurate image labels~\cite{yao2019simulating,yao2022attribute,sun2019dissecting,zhang2021unrealperson,liu2022synthesize}. These methods aim to lower the domain gap between synthetic data and real data by searching a set of parameters that control the 3D rendering process~\cite{yao2019simulating,yao2022attribute}. In comparison, our search is not conducted on predefined parameters but on data directly. This enables more direct research on how to form a good training set. 

\textbf{Object re-ID} has received increasing attention in the past few
years, and many effective systems have been proposed~\cite{khorramshahi2019dual,wang2017orientation,tang2019pamtri,zhou2018aware}. In this paper, we study object re-ID datasets rather than algorithms. Depending on the camera condition, location and environment, existing object re-ID datasets usually have their own distinct characteristics or bias~\cite{sun2019dissecting,yao2022attribute,sun2021ranking}. We show details in \S\ref{sec:motivation}. 

\begin{figure*}[t]
\centering
\includegraphics[width=\linewidth]{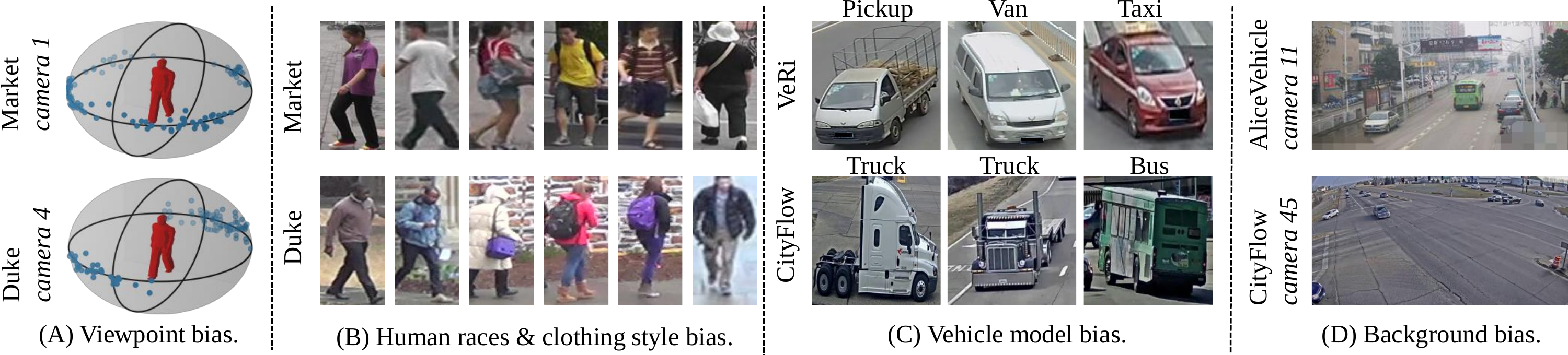}
\caption{Dataset bias in existing object re-ID datasets. (A) Viewpoint distribution of different re-ID datasets. Each blue dot indicates a sample filmed from a specific viewpoint. Compared with the bi-modal viewpoint distribution in Market camera 1, Duke camera 4 has more diverse viewpoints. (B) Different human races and clothing styles of different person re-ID datasets. (C) Different vehicle makes in different datasets. For example, taxis in VeRi are unlikely to be found in CityFlow. (D) Different backgrounds in different datasets. We show a city background in AliceVehicle versus an urban background in CityFlow. To tackle such bias in the target domain, we design an automatic way to generate a training set with similar distribution or bias. }
\label{fig:data_bias}
\vspace{-1em}
\end{figure*}

\section{Motivation: Tackling Target Bias}
\label{sec:motivation}
Data bias commonly exists in the re-ID datasets (for examples see below)~\cite{zhang2021unrealperson,wang2020surpassing,yan2017exploiting,lou2019veri}, and it becomes problematic when the training and testing have different biases. Given a target domain with a certain bias, we aim to find a target-specific training set that has similar distribution or bias. Depending on the filming scenario, there are four major types of data biases in existing re-ID datasets. We show examples of each type in Fig.~\ref{fig:data_bias}.

{\bf Viewpoint.} Viewpoint bias applies to generic objects, including persons and vehicles.
We visualize the viewpoint distributions of two representative person re-ID datasets in Fig.~\ref{fig:data_bias}(A), \ie,~Market-1501~\cite{zheng2015scalable} (denoted as Market) and Duke-reID~\cite{zheng2017unlabeled} (denoted as Duke)\footnote{We understand that it is no longer encouraged to use the Duke dataset. In fact, we are not using it individually for algorithm design, but moving forward to find solutions to replace such individual use.}.

{\bf Human races and clothing style.} Subject to the places where the data is collected, identities in the person re-ID datasets can have distinctive patterns. We show in Fig.~\ref{fig:data_bias}(B) that the humans in Market and Duke datasets evince different races and clothing styles. 

{\bf Vehicle model.} Similar to the identity bias~(\ie, human races and clothing style) in person re-ID, the vehicle identities in vehicle re-ID also hold distinct patterns across different datasets. We show examples in Fig.~\ref{fig:data_bias}(C) using the VeRi~{\cite{liu2016large}} (denoted as VeRi) and the CityFlow~\cite{tang2019cityflow}.

{\bf Background.} Background bias exists in both person and vehicle re-ID datasets, which is similar to viewpoint bias. Fig.~\ref{fig:data_bias}(D) compares the background difference of images from two different vehicle re-ID datasets.

\section{Method}

Given a target unlabeled dataset, we aim to construct a source labeled dataset that has minimal data bias inconsistencies with the target, under certain budget constraints. It induces the model trained on the source to show good performance on the target. To achieve the construction of the source (training) dataset, we propose the search and pruning (SnP) framework. 

\subsection{Overview}
We denote the target set as $\mathcal{D}_T=\{({\ve x}_i,y_i)\}_{i\in[m_t]}$ 
where $m_t$ indicates the number of image-label pairs in the target and $[m_t]=\{1,2,\dots,m_t\}$. It follows the distribution $p_T$, \ie,~$\mathcal{D}_T\sim p_T$. Let $\mathcal{D}_S$ be the source set to be constructed under a budget~$b$. The budget is specified by the number of identities $n$ and the number of images $m$ allowed in $\mathcal{D}_S$, denoted as $b=(n,m)$. A high budget can lead to an unwanted increase in the training cost, in terms of either training time or model size.

To construct the training set $\mathcal{D}_S$, we create a source pool $\mathcal{S}$, which is a collection of multiple object re-ID datasets. It is represented as $\mathcal{S}=\mathcal{D}_S^1\bigcup\mathcal{D}_S^2\dots\bigcup\mathcal{D}_S^K$. Here each $\mathcal{D}_S^k$, $k\in[K]$, indicates the $k$-th source re-ID dataset. Given the source pool, we firstly build a subset ${\mathbf S}^*$ of $\mathcal{S}$ regardless of the budget constraint. Let $h_{\mathbf S}$ be a model $h$ trained on an arbitrary dataset ${\mathbf S}$. The prediction risk of $h_{\mathbf S}$ on the test sample $\ve x$ with ground truth label $y$ is computed as $\ell(h_{\mathbf S}({\ve x}), y)$. We build ${\mathbf S}^*$ by ensuring that the model $h_{{\mathbf S}^*}$ has minimized risk on $\mathcal{D}_T$, \ie, 
\begin{equation}
{\mathbf S}^* = \argmin_{\mathbf S \in 2^\mathcal{S}} \mathbb{E}_{{\boldsymbol x},y \sim p_T}[\ell(h_{\mathbf S}({\ve x}), y)]. 
\label{eq:problem_define}
\end{equation}
We apply target-specific search in \S\ref{subsec:search} to construct ${\mathbf S}^*$.

It can be seen that the construction of ${\mathbf S}^*$ does not take the budget constraint $b=(n,m)$ into consideration, which is otherwise important in reality. Therefore, we build the training set $\mathcal{D}_S$ by pruning ${\mathbf S}^*$ to comprise no more than $n$ identities and no more than $m$ images.  Details of the budgeted pruning process are introduced in \S\ref{subsec:prune}. 

\begin{figure*}[t]
\centering
\includegraphics[width=\linewidth]{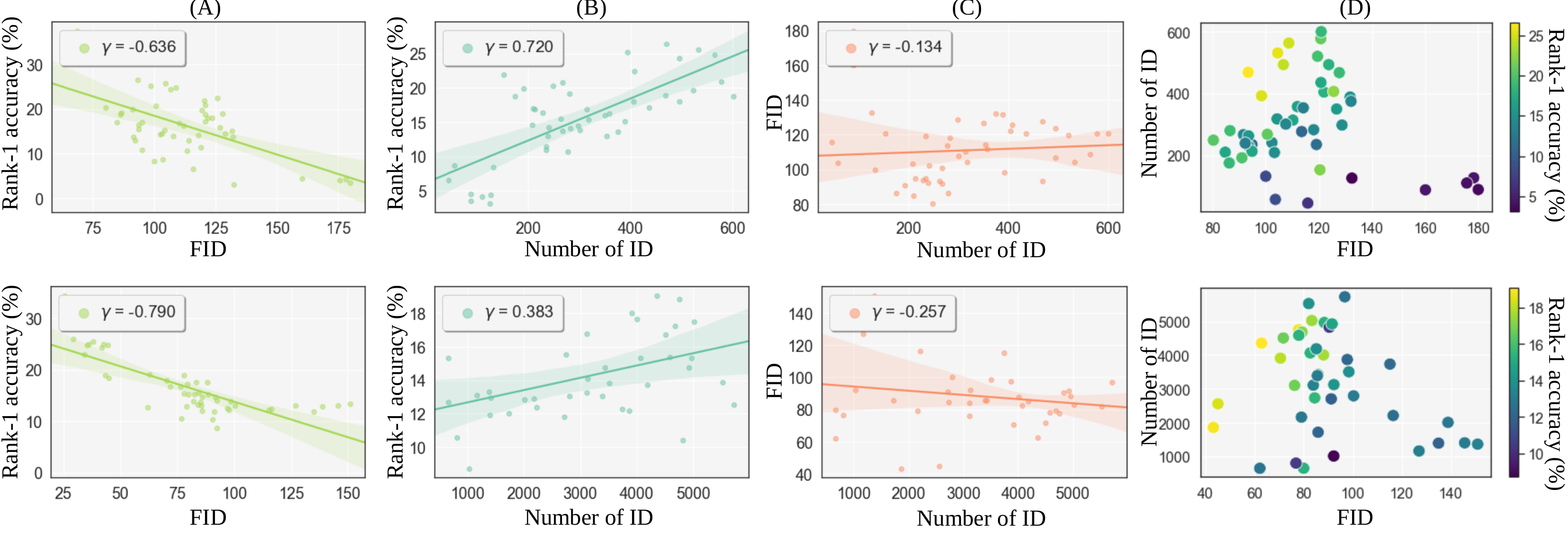}
\caption{ Relationships between the domain gap (measured using FID), the number of ID and the rank-1 accuracy on the target. \textbf{Top}: correlations when AlicePerson~\cite{alice_benchmark} is targeted. \textbf{Bottom}: correlations when AliceVehicle~\cite{alice_benchmark} is targeted. The Pearson correlation is used to measure the relationship between them. (A) FID \vs the rank-1 accuracy. They have a relatively strong negative correlation $(\leq-0.636)$. (B) The number of ID \vs the rank-1 accuracy. There exists a positive correlation but such a correlation is not stable. (C) FID \vs number of ID. The correlation is weak between them. (D) The joint influence of FID and the number of ID on rank-1 accuracy. The top left corner, \ie, training sets that have low FID scores and a large number of IDs have high rank-1 accuracy on the target.
}
\label{fig:correlation_study}
\vspace{-1em}
\end{figure*}

\subsection{Correlation Study}

In order to know how to obtain ${\mathbf S}^*$, we conduct the correlation study to learn the relationships between the dataset bias differences (\ie, domain gap measured in measured in FID~\cite{heusel2017gans}), the number of IDs and the training set quality (\ie, rank-1 test accuracy with the model trained on the such training set). Shown in Fig.~\ref{fig:correlation_study}, in each subfigure, each point represents a training set, which is clustered from the source pool. For each training set, we calculate its domain gap to the target domain, count its number of IDs, and evaluate the rank-1 accuracy. We use the Pearson correlation coefficient ($\gamma $) ~\cite{cohen2009pearson} to measure the correlation between them. The Pearson correlation coefficients range from $[−1, 1]$. A value closer to −1 or 1 indicates a stronger negative or positive correlation, respectively, and 0 implies no correlation. 

From this Fig.~\ref{fig:correlation_study}(A), we observe a relatively strong negative correlation between domain gap and rank-1 accuracy. This indicates minimizing the domain gap between the source and target is highly likely to improve training set quality, \ie, test set rank-1 accuracy. From this Fig.~\ref{fig:correlation_study}(B), we observe a positive but unstable correlation between the number of ID and the rank-1 accuracy. For example, when AlicePerson is targeted, they have a relatively strong positive correlation $(0.720)$. However, such correlation is only $0.383$ when AliceVehicle is targeted. Fig.~\ref{fig:correlation_study}(C) shows the correlation between FID and the number of ID is weak. They are independent factors that influence training set quality. Fig.~\ref{fig:correlation_study}(D) shows the joint influence of FID and the number of ID on the training set quality. The top left corner, \ie, training sets that have a low domain gap to the target and a large number of IDs are of high quality, thereby can train a model that has higher rank-1 accuracy on the target.

\subsection{Target-specific Subset Search}
\label{subsec:search}
The theory of domain adaptation~\cite{ben2010theory} states that:
\begin{equation}
\varepsilon_T(h)<\varepsilon_S(h) + \frac{1}{2}d_{\mathcal{H}\Delta\mathcal{H}}({\mathbf S}, {\mathcal{D}_T}).
\label{eq:bound}
\end{equation}
Here $h\in \mathcal{H}$ represents the hypothesis function (\ie, the model).  $\varepsilon_T(h)$ is  the risk of model $h$ on the target set $\mathcal{D}_T$, while $\varepsilon_S(h)$ is its risk on the source set ${\mathbf S}$.   $d_{\mathcal{H}\Delta\mathcal{H}}({\mathbf S}, {\mathcal{D}_T})$ is the unlabelled $\mathcal{H}\Delta\mathcal{H}$ divergence~\cite{ben2010theory} between ${\mathbf S}$ and ${\mathcal{D}_T}$. Equation~\ref{eq:bound} shows that the target risk $\varepsilon_T(h)$ is upper bounded by $d_{\mathcal{H}\Delta\mathcal{H}}({\mathbf S}, {\mathcal{D}_T})$.

In the common practice of feature-level domain adaptation, the source training set $\mathbf S$ is fixed while the joint feature extraction model is used to minimize $d_{\mathcal{H}\Delta\mathcal{H}}(\mathbf S, \mathcal{D}_T)$~\cite{long2015learning}. In our design, the feature extraction model is fixed instead. 
To minimize $\varepsilon_{T}(h)$, the problem is reformulated as
\begin{equation}
{\mathbf S}^* = \argmin_{{\mathbf S} \in 2^\mathcal{S}}d_{\mathcal{H}\Delta\mathcal{H}}({\mathbf S},{\mathcal{D}}_T). 
\label{eq:problem_define_diverge}
\end{equation}
Generally, $d_{\mathcal{H}\Delta\mathcal{H}}({\mathbf S}^*,\mathcal{D}_T)$ is difficult to compute, but many alternatives exist in the literature~\cite{long2015learning}. We use Fr\'{e}chet Inception Distance (FID)~\cite{heusel2017gans}, which is defined as:
\begin{equation}
\mbox{FID}({\mathbf S}^*, \mathcal{D}_T) = \left \| \bm{\mu}_s - \bm{\mu}_t  \right \|^{2}_{2} + 
         \Tr(\bm{\Sigma}_s + \bm{\Sigma}_t -2 (\bm{\Sigma}_s \bm{\Sigma}_t)^{\frac{1}{2}}).
\label{eq:fid}
\end{equation}
In Eq. \ref{eq:fid}, $\bm{\mu}_s \in \mathbb{R}^d$ and $\bm{\Sigma}_s \in \mathbb{R}^{d\times d}$ are the mean and covariance matrix of the image descriptors of ${\mathbf S}^*$, respectively. $\bm{\mu}_t$ and $\bm{\Sigma}_t$ are those of $\mathcal{D}_T$. $\Tr(.)$ represents the trace of a square matrix. $d$ is the dimension of the image descriptors. Consequently, the objective function is reduced as, 
\begin{equation}
{\mathbf S}^* = \argmin_{{\mathbf S} \in 2^\mathcal{S}}\mbox{FID}({\mathbf S}, \mathcal{D}_T).
\label{eq:problem_define_fid}
\end{equation}
We build ${\mathbf S}^*$ with the greedy algorithm below.

\begin{figure}[t]
\vspace{-1em}
\begin{center}
	\includegraphics[width=1\linewidth]{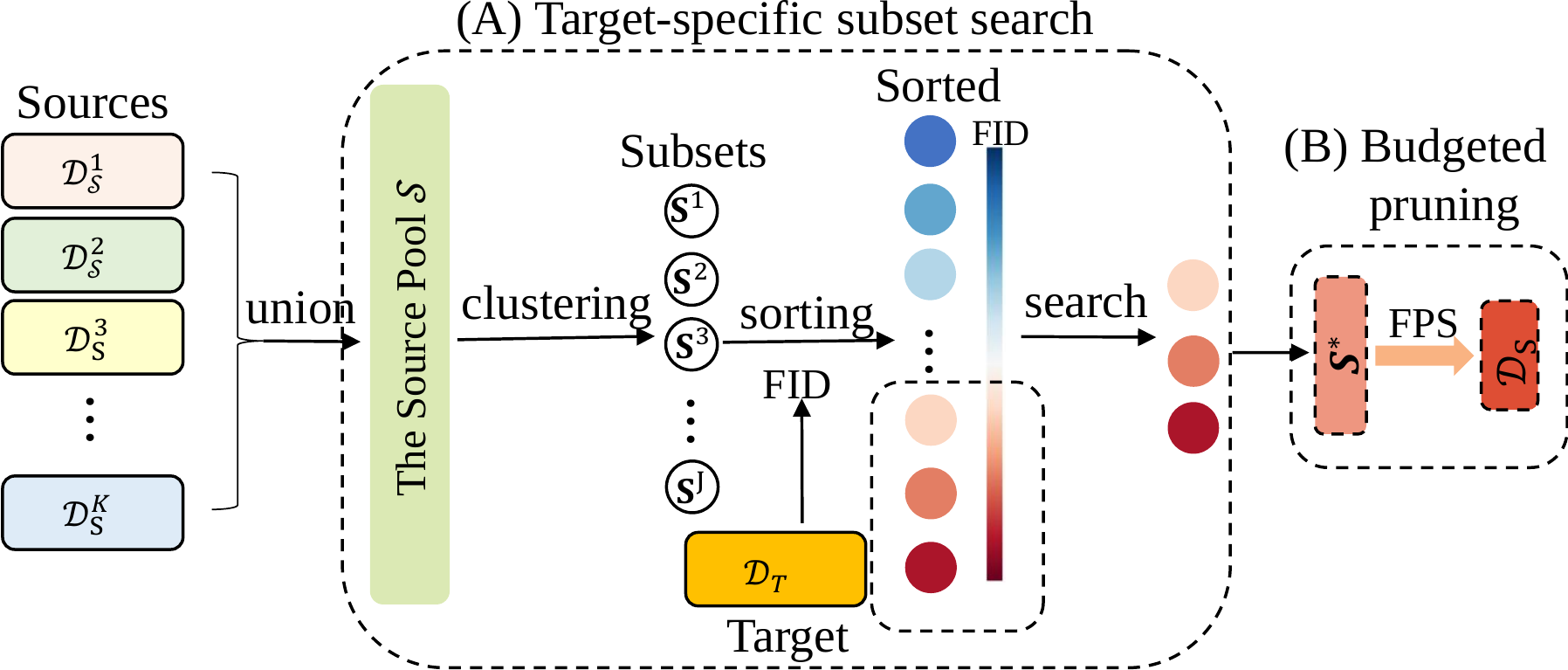}
\end{center}
\vspace{-1em}
\caption{Workflow of the proposed SnP method. We are given sources composed of $K$ existing datasets. From these sources, we aim to construct a training set, which satisfies the budget of no more than $n$ IDs and $m$ images. To achieve this, we perform (A) target-specific subset search to obtain a subset $\mathbf{S}^*$ with similar distributions to the target, then perform (B) budgeted pruning to select $n$ IDs and $m$ representative images, forming the final training set. }
\label{fig:STD-reid-workflow}
\vspace{-1em}
\end{figure}
\begin{figure}[t]
\begin{center}	\includegraphics[width=1\linewidth]{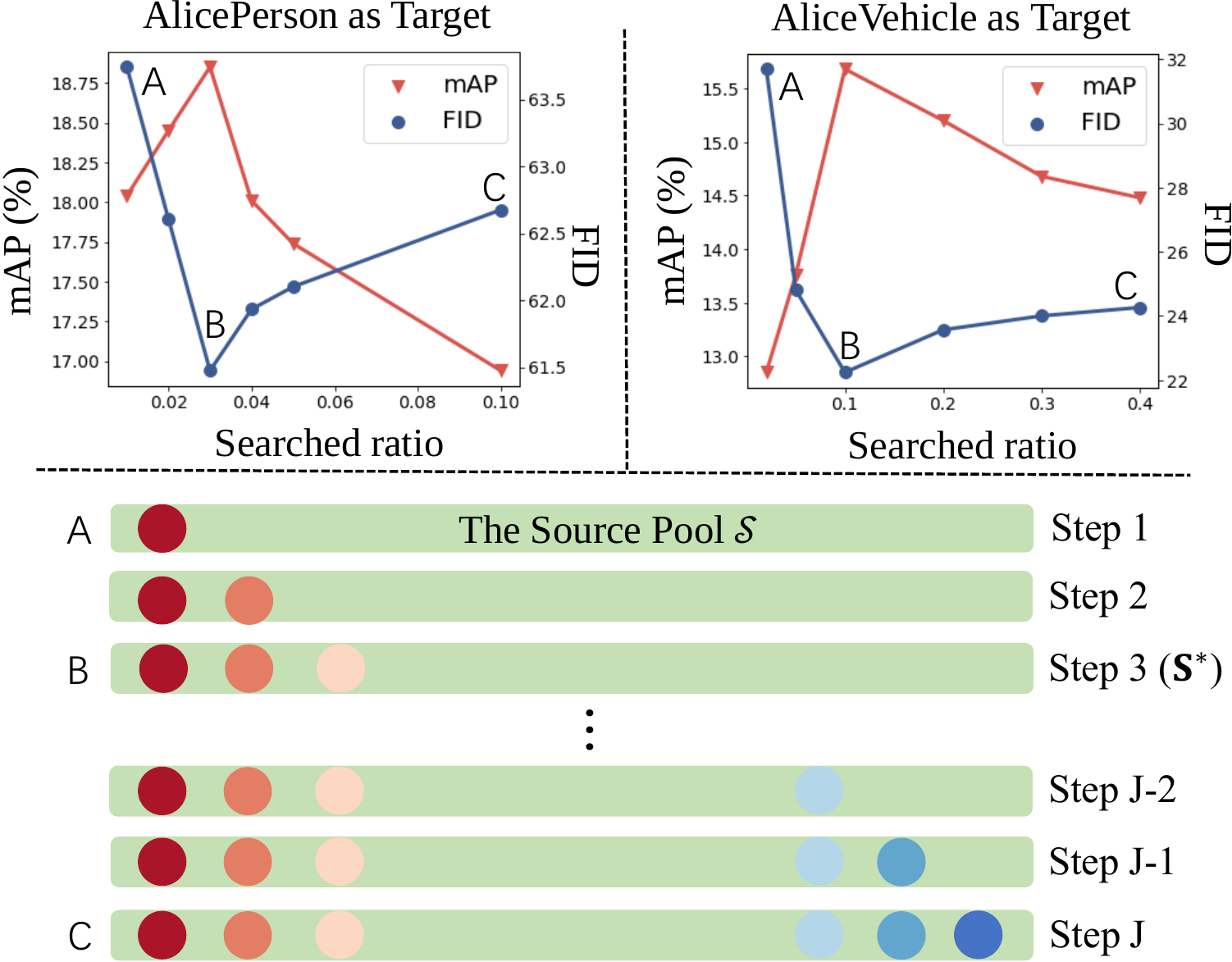}
\end{center}
\caption{Performance of constructed training set at different iterations on the target set. 
We greedily add clustered subset to form ${\mathbf S}^*$ which has the smallest domain gap to the target $\mathcal{D}_T$. During this process, we have relatively high FID and low accuracy at (A). Gradually, we reach situation (B) of the smallest domain gap and highest accuracy. After (B), adding subsets leads to an increase in the domain gap, and the accuracy drops, \eg, (C). To study the correlation between FID and mAP during the search process, we hope to eliminate the impact of dataset size. Thus we use the same number of IDs from the searched set (\ie, 2\% ID). }
\label{fig:domain_adaptive}
\vspace{-1em}
\end{figure}

Firstly, we divide the entire dataset $\mathcal{S}$ into $J$ clusters $\{{\mathbf S}^{1},\cdots,{\mathbf S}^{J} \}$, 
as shown in Fig.~\ref{fig:STD-reid-workflow} (A). Specifically, we average all image descriptors that belong to the same identity, and use this ID-averaged descriptor to represent all corresponding images. Afterwards, we cluster the ID-averaged features into $J$ groups using the k-means method~\cite{likas2003global}. 
Each subset ${\mathbf S}^{j}$, $j\in[J]$, is  composed of all images with the corresponding IDs in that group. 
Secondly, we calculate the FID between each subset ${\mathbf S}^{j}$ and the target $\mathcal{D}_T$, and sort $\{\mbox{FID}({\mathbf S}^{j}, \mathcal{D}_T)\}_{j\in[J]}$ in ascending order. To build ${\mathbf S}^*$, we keep  adding the subsets with lower FID to ${\mathbf S}^*$ until $\mbox{FID}({\mathbf S}^*, \mathcal{D}_T)$ stops to decrease. This indicates that the constructed ${\mathbf S}^*$ holds the minimum FID to the target set. Algorithm~\ref{algorithm:domain_search} summarizes the above procedures. 
We empirically demonstrate in Fig. \ref{fig:domain_adaptive} that, the subset ${\mathbf S}^*$ with the minimum $\mbox{FID}({\mathbf S}^*,\mathcal{D}_T)$ results in the model to produce highest accuracy on the target dataset.

\subsection{Budget-constrained Pruning}
\label{subsec:prune}
Using the target-specific subset search, we construct a candidate training set ${\mathbf S}^*$. Yet, it can violate the budget constraint of $b=(n,m)$, which prefers the training set to include no more than $n$ identities and no more than $m$ images. 
Let $\mathcal{Y}({\mathbf S}^*)=\{y^1,y^2,\dots,y^{a}\}$, be the set of unique identities in ${\mathbf S}^*$, and ${\ve s}(y^i)$, $i\in[a]$ be the set of all images with identity $y^i$. Generally, 
we have $a\geq n$. 
To get the training set $\mathcal{D}_S$ under budget $b$, we randomly sample a subset ${\ve y}$ of $n$ identities from $\mathcal{Y}$. 
All images with identities in ${\ve y}$ are combined to form a subset $\hat{\mathbf s}=\{{\ve s}(y^i)|y^i\in{\ve y}\}$. We also initialize $\mathcal{D}_S$ by random sampling a seed image from each 
${\ve s}(y^i),~y^i\in{\ve y}$. This guarantees $\mathcal{D}_S$ to cover all the identities in the subset $\hat{\mathbf s}$. 


Suppose the number of images in $\hat{\mathbf s}$ is larger than $m$. We construct $\mathcal{D}_S$ by sampling images iteratively from $\hat{\mathbf s}$ until $|\mathcal{D}_S|=m$. To ensure similar performance between the model trained on $D_S$ and the model trained on $\hat{\mathbf s}$, we minimize the risk differences between them, 
\begin{equation}
\mathcal{D}_S = \argmin_{\mathcal{D}_S \in 2^{\hat{\mathbf s}}}\big|L(h_{\hat{\mathbf s}}({\ve x}), y) -
L(h_{\mathcal{D}_S}({\ve x}), y)\big|.
\label{eq:problem_define_pruning}
\end{equation}
$L(h_{\hat{\mathbf s}}({\ve x}), y)$ and $L(h_{\mathcal{D}_S}({\ve x}), y)$ are the respective risks of model $h$ on the dataset ${\hat{\mathbf s}}$ and $\mathcal{D}_S$.
For explicity, we define the risk of model $h$ on an arbitrary dataset $\mathbf{S}$ as 
\begin{equation}
L(h_{{\mathbf S}}({\ve x}), y)=\frac{1}{|{\mathbf S}|} \sum_{({\ve x}_i, y_i) \in {\mathbf S}} \ell(h_{{\mathbf S}}({\ve x}_i), y_i).
\label{eq:average_loss}
\end{equation}
As in Eq.~\ref{eq:problem_define}, $\ell(h_{{\mathbf S}}({\ve x}_i), y_i)$ is the risk on individual samples. 

From the theory of core set~\cite{sener2018active}, if $\mathcal{D}_S$ is the $\delta_s$ cover of the set $\hat{\mathbf s}$ and shares the same number of classes with $\hat{\mathbf s}$, the risk difference between model $h_{\hat{\mathbf s}}$ and $h_{\mathcal{D}_S}$ is bounded by
\begin{equation}
\big|L(h_{\hat{\mathbf s}}({\ve x}), y) - L(h_{\mathcal{D}_S}({\ve x}), y)\big| \leq \mathcal{O}(\delta_s) + \mathcal{O}(\frac{1}{\sqrt{|\mathcal{D}_S|}}).
\label{eq:coreset}
\end{equation}
\begin{algorithm}[t]  
  \caption{Target-specific Subset Search}  
  \begin{algorithmic}[1]
    \State \textbf{Input:} Source pool $\mathcal{S}$, target set $\mathcal{D}_T$, and the number of clusters $J$. 
    \State \textbf{Begin:}
    \State Cluster ($\mathcal{S}, J)\longrightarrow\{{\mathbf S}^1, \cdots, {\mathbf S}^J \}$ 
    \State ${\mathbf s}=\emptyset, \epsilon=\infty, {\mathbf S}^*=\emptyset$  
    \State $\mathcal{J}=\mbox{argsort}\big(\{\mbox{FID} ({\mathbf S}^j, {\mathcal{D}_T})\}_{j\in[J]}\big)$ \Comment{Ascending}
    \For {$j$ in $\mathcal{J}$}
         \State ${\mathbf s}={\mathbf s}\mathop{\cup}{\mathbf S}^j$
        \If {$\mbox{FID}({\mathbf s}, T)<\epsilon$}
            \State $\epsilon=\mbox{FID} ({\mathbf s}, {\mathcal D}_T)$ 
            \State ${\mathbf S}^*={\mathbf s}$ 
        \EndIf 
    \EndFor
  \State \Return ${\mathbf S}^*$
  \end{algorithmic}  
  \label{algorithm:domain_search}

\end{algorithm} 
$\delta_s$ is the radius of the cover, and $\mathcal{O}(\delta_s)$ is a polynomial function over $\delta_s$. The problem can be reduced as a K-center problem~\cite{farahani2009facility} by optimizing $\mathcal{O}(\delta_s)$. 
We apply a 2-approximation algorithm \cite{williamson2011design} to iteratively find optimal samples in ${\hat{\mathbf s}}$ and add to $\mathcal{D}_S$. Specifically,  each optimal sample ${\mathbf z}^*$ is computed as
\begin{equation}
{\mathbf z}^* = \argmax\limits_{{\mathbf z}_i \in {\hat{\mathbf s}} \setminus \mathcal{D}_S}\min_{{\mathbf z}_j \in \mathcal{D}_S} \|f({\ve x}_i)- f({\ve x}_j)\|_2, 
\label{eq:FPS}
\end{equation}
where ${\mathbf z}=({\ve x},y)$, and $f({\ve x})$ represents the descriptor of an image ${\ve x}$. Equation~\ref{eq:FPS} relates to the furthest point sampling method~\cite{eldar1997farthest}, which enables the most representative samples from a dataset to be selected. We summarize the prunning process in Algorithm~\ref{algorithm:budget_pruning}.

\subsection{Discussion}
\textbf{What to prune first, ID or image?} In our design, we select the IDs first and then images, aiming to build a dataset that has a small domain gap to target and is meanwhile in small-scale. Selecting the images first and then IDs is possible. However, it leads to a significant increase in the time complexity. As shown in Algorithm \ref{algorithm:budget_pruning}, the most time-consuming part of our algorithm is the FPS for finding image core sets, which takes $\mathcal{O}(m|\hat{\mathbf s}|)$. If we select images first, the time complexity becomes $\mathcal{O}(m|{\mathbf S}^*|)$, which is significantly higher than $\mathcal{O}(m|\hat{\mathbf s}|)$ as $|{\mathbf S}^*| \gg |\hat{\mathbf s}|$.

\textbf{Is SnP applicable to tasks beyond object re-ID?} It is possible to apply our search and pruning (SnP) framework to other tasks, \eg, classification. 
Yet, achieving this requires the SnP algorithm
to be partially redesigned. The reason is that, in object re-ID, the major domain gap comes from the class difference, \ie., IDs. However, such a gap does not exist in the classification tasks as the train/val/test datasets always share the same classes. Therefore, our current design of selecting similar IDs for similar distributions is not directly applicable, and has to be adapted. 




\section{Experiment}

We evaluate the effectiveness of SnP on person re-ID and vehicle re-ID. In both tasks, given target data, we use the SnP pipeline to find a training set that has a similar distribution to the target and simultaneously meets a budget.

\subsection{Source and Target Datasets}
\textbf{Person re-ID.} We create the source pool for person re-ID using $10$ public datasets, including Market~\cite{zheng2015scalable}, Duke~\cite{zheng2017unlabeled}, MSMT17~\cite{wei2018person} (denoted as MSMT), CUHK03~\cite{li2014deepreid}, RAiD~\cite{das2014consistent}, PersonX~\cite{sun2019dissecting}, UnrealPerson~\cite{zhang2021unrealperson}, RandPerson~\cite{wang2020surpassing}, PKU-Reid~\cite{ma2016orientation} and VIPeR~\cite{cho2018pamm}. Those datasets cover both synthetic and real-world data. Ten re-ID datasets constitute a source pool that contains in total of 15,060 IDs and 399,715 images. 

We use two real-world datasets as targets: AlicePerson~\cite{alice_benchmark} and Market~\cite{zheng2015scalable}. Specifically, AlicePerson is specially designed for domain adaptation as it contains unlabeled training images. 
Note that in Market, the label for its training set is not be used. When Market is the target, Market training set is excluded from the source pool. 

\begin{algorithm}[t]  
  \caption{Budgeted Pruning}  
  \begin{algorithmic}[1]
    \State \textbf{Input:} Initial source set ${\mathbf S}^*$, budget $b=(n,m)$. 
    \State \textbf{Begin:}
    \State ${\ve y}=\mathcal{U} (\mathcal{Y}({\mathbf S}^*),n)$ \Comment{Sample $n$ IDs}
    \State $\hat{\mathbf s}=\{{\ve s}(y_i)|y_i\in{\ve y}\}$ 
    \State $\mathcal{D}_S=\{\mathcal{U}({\mathbf s}(y^i),1)|y_i\in{\ve y}\}$ 
  \If {$|\hat{\mathbf s}| > m $}  \Comment{Sample $m$ images}
        \Repeat          
        \State ${\mathbf z}^* = \argmax\limits_{{\mathbf z}_i \in \hat{\mathbf s} \setminus \mathcal{D}_S}\min\limits_{{\mathbf z}_j \in \mathcal{D}_S} \|{\ve x}_i - {\ve x}_j\|_2$
        \State $\mathcal{D}_S = \mathcal{D}_S \cup \{{\mathbf z}^*\} $ 
    \Until{ $|\mathcal{D}_S| = m $}
    \Else
    \State $\mathcal{D}_S=\hat{\mathbf s}$
  \EndIf
  \State \Return $\mathcal{D}_S$
  \end{algorithmic}  
\label{algorithm:budget_pruning}
\end{algorithm}

\textbf{Vehicle re-ID.} For vehicle re-ID, we create the source pool by combining $8$ datasets, which are VeRi~\cite{liu2016large}, CityFlow~\cite{tang2019cityflow}, VehicleID~\cite{liu2016deep}, VeRi-Wild~\cite{lou2019veri}, VehicleX~\cite{yao2019simulating}, Stanford Cars~\cite{krause20133d}, PKU-vd1~\cite{yan2017exploiting} and PKU-vd2~\cite{yan2017exploiting}. It totally has 156,512 IDs and 1,284,272 images. 

AliceVehicle~\cite{alice_benchmark} and VeRi~\cite{liu2016large} are separately used as the target domains. Similar to AlicePerson, AliceVehicle is also designed for domain adaptation. 
When VeRi is the target, the VeRi training set is excluded from the source pool.

\textbf{Evaluation protocol.} For object re-ID, we use mean average precision (mAP) and cumulative match curve (CMC) scores to measure system accuracy, \eg, ``rank-1'' and ``rank-5''. ``rank-1'' denotes the success rate of finding the true match in the first rank, and ``rank-5'' means the success rate of ranking at least one true match within top-5 matches.

\begin{table*}[t]
\centering
\footnotesize
\caption{Comparing different methods in training data search: SnP, random sampling, and greedy sampling. We set the budget as $2\%$, $5\%$, and $20\%$ of the total source IDs. We use four targets: AlicePerson, Market, AliceVehicle and VeRi. The task model is IDE~\cite{zheng2016mars}. FID, rank-1 accuracy (\%), and mAP (\%) are reported. 
}
\setlength{\tabcolsep}{2.25mm}
\begin{tabular}{ccl|cccccc|cccccc}
\Xhline{1.2pt}
\multicolumn{3}{c|}{\multirow{3}{*}{Training data}}                                             & \multicolumn{6}{c|}{Person re-ID targets}                                        & \multicolumn{6}{c}{Vehicle re-ID targets}                                     \\ \cline{4-15} 
\multicolumn{3}{l|}{}                                                                           & \multicolumn{3}{c|}{AlicePerson}          & \multicolumn{3}{c|}{Market} & \multicolumn{3}{c|}{AliceVehicle}         & \multicolumn{3}{c}{VeRi} \\ \cline{4-15} 
\multicolumn{3}{l|}{}                                                                           & FID$\downarrow$    & R1$\uparrow$    & \multicolumn{1}{l|}{mAP$\uparrow$}   & FID$\downarrow$      & R1$\uparrow$      & mAP$\uparrow$    & FID$\downarrow$    & R1$\uparrow$   & \multicolumn{1}{l|}{mAP$\uparrow$}   & FID$\downarrow$     & R1$\uparrow$    & mAP$\uparrow$    \\ \hline
\multicolumn{3}{c|}{Source pool}                                                               & 81.67 & 38.96 & \multicolumn{1}{l|}{17.62} & 37.53   & 55.55   & 30.62   & 43.95 & 30.47 & \multicolumn{1}{l|}{14.64} & 24.39  & 55.90  & 25.03  \\ \hline
\multicolumn{3}{c|}{Searched}                                                                & 60.95 & 48.19 & \multicolumn{1}{l|}{25.51} & 30.42   & 61.49   & 34.40   & 23.44 & 46.78 & \multicolumn{1}{l|}{25.46} & 17.92  & 74.13  & 41.71  \\ \hline
\multicolumn{1}{l|}{\multirow{9}{*}{\rotatebox{90}{Pruned}}} & \multicolumn{1}{l|}{}         & Random  & 80.06 & 23.67 & \multicolumn{1}{l|}{9.80}   & 39.03   & 40.77   & 19.54   & 45.97 & 31.35 & \multicolumn{1}{l|}{12.87} & 24.68  & 67.16  & 26.48  \\
\multicolumn{1}{l|}{}                         & \multicolumn{1}{l|}{2\% IDs}  & Greedy ~\cite{yan2020neural} & 61.78 & 33.22 & \multicolumn{1}{l|}{15.10}  & 34.43   & 42.19   & 19.35   & 38.51 & 31.82 & \multicolumn{1}{l|}{12.91} & 29.53  & 66.57  & 26.24  \\
\multicolumn{1}{l|}{}                         & \multicolumn{1}{l|}{}         & SnP             & 60.42 & 38.23 & \multicolumn{1}{l|}{18.17} & 31.29    & 44.80    & 21.65   & 23.48 & 38.48 & \multicolumn{1}{l|}{17.79} & 17.70  & 69.96  & 31.10   \\ \cline{2-15} 
\multicolumn{1}{l|}{}                         & \multicolumn{1}{l|}{}         & Random    & 81.41 & 33.16 & \multicolumn{1}{l|}{14.49} & 39.65   & 47.39   & 23.97   & 44.52 & 36.36  & \multicolumn{1}{l|}{14.17} & 25.27  & 70.38   & 30.44  \\
\multicolumn{1}{l|}{}                         & \multicolumn{1}{l|}{5\% IDs}  & Greedy ~\cite{yan2020neural} & 61.01 & 44.63  & \multicolumn{1}{l|}{22.81} & 31.63   & 49.17   & 24.77   & 32.48 & 41.32 & \multicolumn{1}{l|}{17.77} & 26.06  & 71.23  & 32.53  \\
\multicolumn{1}{l|}{}                         & \multicolumn{1}{l|}{}         & SnP             & 60.64  &47.26  & \multicolumn{1}{l|}{25.45} & 30.37   & 51.96   & 26.56   & 23.92 & 44.58 & \multicolumn{1}{l|}{21.79} & 18.09  & 72.05  & 36.01  \\ \cline{2-15} 
\multicolumn{1}{l|}{}                         & \multicolumn{1}{l|}{}         & Random  & 79.33  & 38.10  & \multicolumn{1}{l|}{17.79} & 38.63   & 53.15   & 28.39   & 43.90 & 40.89  & \multicolumn{1}{l|}{18.13} & 24.43  & 68.71  & 34.10   \\
\multicolumn{1}{l|}{}                         & \multicolumn{1}{l|}{20\% IDs} & Greedy ~\cite{yan2020neural} & 63.15 & 46.74 & \multicolumn{1}{l|}{22.65} & 32.42   & 53.53   & 28.19   & 24.15 & 44.58 & \multicolumn{1}{l|}{22.82} & 18.74  & 71.04  & 38.07  \\
\multicolumn{1}{l|}{}                         & \multicolumn{1}{l|}{}         & SnP             & 61.87  & 47.20  & \multicolumn{1}{l|}{25.36} & 30.58   & 57.14   & 33.09   & 23.47 & 46.07  & \multicolumn{1}{l|}{25.24} & 17.93  & 73.48  & 40.75  \\ 
\Xhline{1.2pt}
\end{tabular}
\label{tab:snp_to_othermethods}
\vspace{-1em}
\end{table*}

\begin{table}[t]
\centering
\footnotesize
\setlength{\tabcolsep}{0.8mm}
\caption{The effectiveness of the target-specific subset search. mAP (\%) and CMC scores are reported. ``R1'' and ``R5'' denote rank-1 accuracy (\%) and rank-5 accuracy (\%), respectively.}
\begin{tabular}{l|ccc|l|ccc}
\Xhline{1.2pt}
                 & \multicolumn{3}{c|}{AlicePerson } &                 & \multicolumn{3}{c}{AliceVehicle } \\ \hline
Training data   & R1           & R5           & mAP          & Training data   & R1           & R5           & mAP          \\ \hline
Market~\cite{zheng2015scalable}          & 32.89        & 52.54        & 16.06        & VeRi~\cite{liu2016large}            & 30.69        & 43.66        & 11.05        \\
Duke~\cite{zheng2017unlabeled}            & 23.27        & 41.13        & 8.59         & CityFlow~\cite{tang2019cityflow}        & 23.95        & 36.00           & 7.45         \\
MSMT~\cite{wei2018person}           & 31.29        & 52.79        & 15.08        & VehicleID~\cite{liu2016deep}       & 16.3         & 29.13        & 4.73         \\
PersonX~\cite{sun2019dissecting}         & 16.08        & 27.42        & 6.41         & VehicleX~\cite{yao2019simulating}        & 18.85        & 32.18        & 8.89         \\ \hline
Source pool       & 38.96        & 57.48        & 17.62        & Source pool       & 30.47        & 48.33        & 14.64        \\
Searched & 48.19        & 67.57        & 25.51        & Searched & 46.78        & 64.41        & 25.46        \\ \Xhline{1.2pt}
\end{tabular}
\label{tab:search_effectiveness}
\vspace{-1em}
\end{table}


\subsection{Results}
\label{sec:Quantitative_Evaluation}

\textbf{SnP framework \emph{vs.} random sampling and greedy sampling.} Given a target domain, the SnP framework allows us to construct a budgeted dataset with a similar distribution. We show the superiority of SnP framework in Table \ref{tab:snp_to_othermethods}. After sampling target-specific data with SnP, we train the subsequent re-ID model with the sampled target-specific data only. In Table \ref{tab:snp_to_othermethods}, we compare the sampled dataset with those created by greedy sampling and random sampling. Random sampling means we randomly select IDs according to the uniform distribution. For greedy sampling, we reproduce \cite{yan2020neural}. Specifically, We assign each ID a score, which is calculated using FID. IDs will then be selected greedily from the lowest FID value to the highest FID value.

From the results shown in Table \ref{tab:snp_to_othermethods} and Table \ref{tab:sota_models}, we observe that training sets selected using SnP  achieve consistently lower domain gap to the target and higher re-ID accuracy than those found by random sampling and greedy sampling. For example, when creating a training set with only 2\% source IDs for AlicePerson as the target, SnP results in -19.64 and -1.36 improvement in FID value, and +14.56\% and +5.01\% improvement in rank-1 accuracy over using random sampling and greedy sampling, respectively. When sampling training set with 2\% of the source IDs for AliceVehicle as the target, the rank-1 accuracy improvement is +7.13\% and +6.66\%, respectively.

\begin{table}[t]
\centering
\footnotesize
\caption{The superiority of SnP over random sampling and greedy sampling, when different direct transfer and pseudu-label re-ID models are used. We report accuracy when $2\%$ IDs are selected for target AlicerPerson. Notations and evaluation metrics are the same as those in the previous table.  }
\setlength{\tabcolsep}{1.6mm}
\begin{tabular}{c|c|cc|ccc}
\Xhline{1.2pt}
Type                             & Model                      & \multicolumn{2}{l|}{ Training data} & R1 & R5 & mAP \\ \hline
\multirow{9}{*}{ \begin{tabular}[c]{@{}l@{}}Direct\\Transfer\end{tabular}  } & \multirow{3}{*}{IDE~\cite{zheng2016mars}}       & \multicolumn{2}{l|}{Random }   &   23.67 &  42.32  &  9.80   \\
                                 &                            & \multicolumn{2}{l|}{Greedy~\cite{yan2020neural} }   &  33.22  &  54.45  &  15.10 \\
                                 &                            & \multicolumn{2}{l|}{SnP}             &  38.23  & 58.40   &  18.17   \\ \cline{2-7} 
                                 & \multirow{3}{*}{PCB~\cite{sun2018beyond}}       & \multicolumn{2}{l|}{Random }   &   24.79 &  41.60  &   9.91  \\
                                 &                            & \multicolumn{2}{l|}{Greedy~\cite{yan2020neural} }   &  29.07  & 46.01   &   12.59  \\
                                 &                            & \multicolumn{2}{l|}{SnP}             &  32.43  &  49.90  &  15.18   \\ \cline{2-7} 
                                 & \multirow{3}{*}{TransReid~\cite{he2021transreid} } & \multicolumn{2}{l|}{Random }   & 52.04   &  69.73  &   28.47  \\
                                 &                            & \multicolumn{2}{l|}{Greedy~\cite{yan2020neural} }   & 63.73   &  80.24   &  41.88    \\
                                 &                            & \multicolumn{2}{l|}{SnP}             &  64.31  &   80.46 &   42.74  \\ \hline
\multirow{6}{*}{\begin{tabular}[c]{@{}l@{}}Pseudo\\-labeling\end{tabular}}    & \multirow{3}{*}{UDA~\cite{song2020unsupervised}}       & \multicolumn{2}{l|}{Random }   &  32.17  & 54.67   &  15.32   \\
                                 &                            & \multicolumn{2}{l|}{Greedy~\cite{yan2020neural} }   & 36.47   & 52.06   & 17.34    \\
                                 &                            & \multicolumn{2}{l|}{SnP}             &   41.41 & 55.74   &  20.47   \\ \cline{2-7} 
                                 & \multirow{3}{*}{MMT~\cite{ge2020mutual}}       & \multicolumn{2}{l|}{Random }   &   35.94  & 51.91   &  17.25   \\
                                 &                            & \multicolumn{2}{l|}{Greedy~\cite{yan2020neural} }   &  38.64  &  56.48  &   21.18  \\
                                 &                            & \multicolumn{2}{l|}{SnP}             &  43.36  &    60.38& 23.34    \\ \Xhline{1.2pt}
\end{tabular}
\label{tab:sota_models}
\vspace{-1em}
\end{table}

\begin{figure*}[t]
\centering
\includegraphics[width=\linewidth]{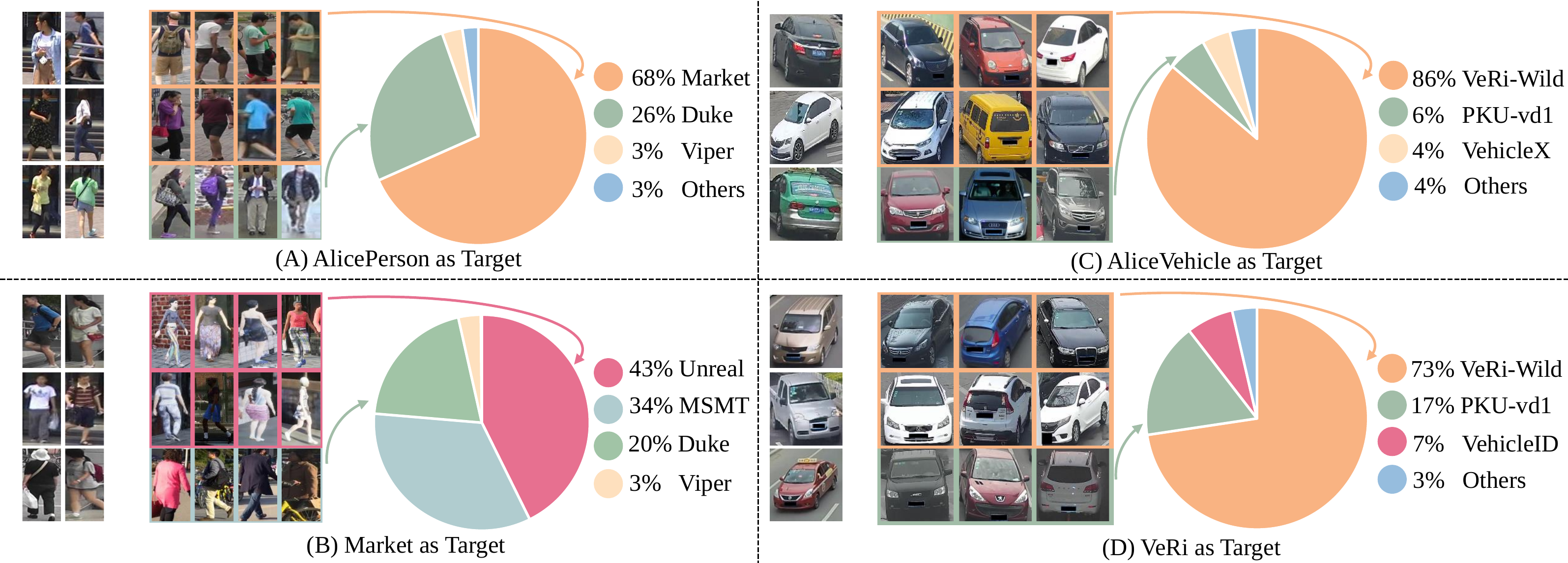}
\caption{Composition statistics and images samples of searched training sets. For each subfigure (A), (B), (C) and (D), the columns on the left presents unlabeled target samples; columns the middle provide samples of the searched training set; the pie chart
Community-verified icon Verified
on the right shows composition statistics of the searched training set. }
\label{fig:pie_chart}
\end{figure*}

\begin{figure*}[t]
\centering
\includegraphics[width=\linewidth]{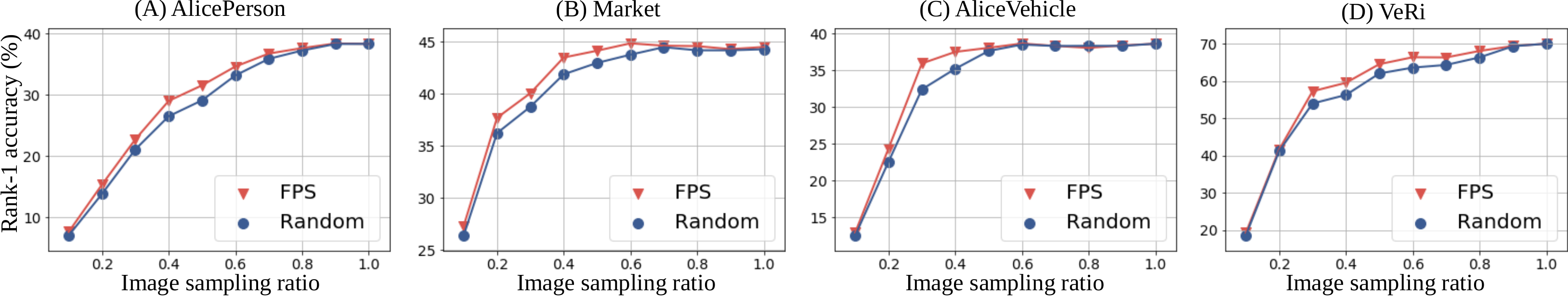}
\caption{Comparing FPS with random sampling. The search step provides us with 2\% of the source IDs. We further select various ratios of images from these IDs. Four different targets are used, and rank-1 accuracy is the evaluation metric. Score is averaged over three runs. }
\label{fig:fps_vs_random}
\vspace{-1em}
\end{figure*}

\textbf{Effectiveness of the search step.} We analyze this step in Table \ref{tab:snp_to_othermethods} and Table \ref{tab:search_effectiveness}. In this search process that poses no limit on the number of IDs and images, we aim to identify and merge clusters of source identities that exhibit similar distributions with the target. We show the searched datasets contribute to improved accuracy over the entire data pool. For example, for AlicePerson, we observe a +9.23\% improvement in rank-1 accuracy over the entire data pool. We further show the searched data is superior to individual training sets in Table \ref{tab:search_effectiveness}. For example, when AliceVehicle is our target, we show our searched dataset results in +16.09\%, +22.83\%, +30.48\%, +27.93\% higher rank-1 accuracy than VeRi, CityFlow, VehicleID, and VehicleX, respectively.

Of note, accuracy under this application scenario is usually lower than that produced by in-distribution training sets. This difference is understandable, because searched data have a relatively lower resemblance to the target data compared with in-distribution training sets. That said, annotating in-distribution training sets is usually expensive, especially considering the complex, specific and ever-changing target environments, where creating a training set on-the-fly with good performance is of practical value.


\textbf{Effectiveness of the pruning step} is analyzed in Table \ref{tab:snp_to_othermethods} and Fig.~\ref{fig:fps_vs_random}. Pruning aims to find a subset that has no more than $n$ IDs and $m$ images. From Table \ref{tab:snp_to_othermethods} and Fig. \ref{fig:fps_vs_random}, admittedly, the pruning of both IDs and images will lead to an accuracy decrease. For example, when the target is AlicePerson, if we select 2\% IDs, there is a -9.23\% decrease in rank-1 accuracy. From Fig. \ref{fig:fps_vs_random}, when the target is AlicePerson, if we further select 40\% images, there is a -9.96\% decrease in rank-1 accuracy. However, even though only 2\% IDs are used, rank-1 accuracy obtained from the pruned training data is still competitive over the source pool, which is just -0.73\% lower than the source pool. It shows the pruning method significantly reduces the training set scale while being able to train a model of reasonable accuracy.   

\textbf{Composition of searched training sets.} We visualize four examples in in Fig.~\ref{fig:pie_chart}. It is clear that searched training sets have different compositions under different targets. 
If we use AlicePerson as the target, in the searched training set, 
images from Market and Duke take up 68\% and 26\%, respectively. In comparison, when using Market as the target, the resulting training set contains, 43\% images from UnrealPerson, and 34\% from MSMT. An interesting observation is that synthetic data (UnrealPerson) has a major role under Market as the target. It demonstrates the potential use of synthetic data for real-world target domains.

\textbf{Comparison between FPS and random sampling.} Both can be used for sampling images resulting from the search step. In Fig. \ref{fig:fps_vs_random}, we sample different ratios of the images resulting from the search step.  
We observe FPS is consistently superior to random sampling, under different selection ratios and targets. 
Improvement of FPS over random sampling first increase and then decreases, with peak improvement happening at the 30-60\% ratio. 

\section{Conclusion}

This paper studies the training data set search problem, for object re-ID applications. Under a certain budget, we aim to find a target-specific training set that gives a competitive re-ID model. We show our method is overall superior to existing strategies such as random sampling and greedy sampling in terms of accuracy on the target domain. We analyze various components in the SnP system and find them to be stable under various source pools and targets. We also point out the correlation between domain gap, dataset size, and training set quality, and would like to further study the data-centric problems in the community.

\section{Acknowledgement}

This work was supported in part by the ARC Discovery Early Career Researcher Award (DE200101283), the ARC Discovery Project (DP210102801), Oracle Cloud credits, and related resources provided by Oracle for Research. 


\section{Experimental Details}

\textbf{SnP settings.} 
For a dataset denoted as a target domain (\eg, Market and VeRi), their unlabeled training sets are used as the search target.
For image feature extraction, we use IncepetionV3~\cite{szegedy2016rethinking} pretrained on ImageNet~\cite{deng2009imagenet}. 
 

\textbf{Task model configuration.} We have two types of models: direct transfer models and pseudo-label based models. For direct transfer, we use multiple task models, including ID-discriminative embedding (IDE)~\cite{zheng2016mars}, the part-based convolution (PCB)~\cite{sun2018beyond}, and TransReid~\cite{he2021transreid}. For IDE, we adopt the strategy from~\cite{luo2019bag} which uses ResNet-50~\cite{he2016deep}, adds batch normalization and removes ReLU after the final feature layer. For PCB, we use the ResNet-50 backbone and vertically partition an image into six equal horizontal parts. For pseudo-label methods, we use UDA~\cite{song2020unsupervised} and MMT~\cite{ge2020mutual}. The unlabeled target training set (\eg, AlicePerson training set) is used for pseudo-label model training. Note that when implementing these task models, we use their official implementation with default hyperparameters including learning rate and training epochs.

\textbf{Computation Resources.} We run experiments on an 8-GPU machine and an oracle cloud machine. The 8-GPU machine has 8 NVIDIA 3090 GPUs and its CPU is AMD EPYC 7343 Processor. For the oracle cloud machine, it has 2 NVIDIA A10 GPUs and its CPU is Intel Platinum 8358 Processor.  For both the 3090 machine and the oracle cloud machine, we use PyTorch version 1.12.1+cu116.

\begin{table}[t]
\centering
\footnotesize
\caption{Generalization comparison of various training sets: those generated by random sampling, greedy sampling, and SnP, as well as the source data pool. AlicePerson is used as the target. We directly test the trained models on Market, Duke, MSMT and report the individual and the average rank-1 accuracy (\%) and mAP (\%).} 
\setlength{\tabcolsep}{0.4mm}
\begin{tabular}{l|cccccccc|cc}
\Xhline{1.2pt}
\multirow{2}{*}{\begin{tabular}[c]{@{}l@{}}Train. set\end{tabular}} & \multicolumn{2}{c}{Market} & \multicolumn{2}{c}{Duke} & \multicolumn{2}{c}{MSMT} & \multicolumn{2}{c|}{Average} & \multicolumn{2}{l}{AlicePerson} \\ \cline{2-11} 
                               & R1          & mAP          & R1         & mAP         & R1         & mAP         & R1           & mAP           & R1             & mAP            \\ \hline
Sour. pool                         &   67.36          &         43.16    &      62.39      &     41.15        &   35.72         &     15.40       &      55.16        &       33.23        &          38.96     &            17.62   \\
Random                         &      6.44       &    1.92          &    38.29        &    18.75         &       19.90     &     6.42        &       21.54       &       9.03        &       23.67         &     9.80           \\
Greedy                         &      74.11       &      51.98        &      26.08      &    13.27         &       7.32     &     2.13        &     35.83        &        22.46       &      33.22          &    15.10            \\
SnP                            &       74.32      &      51.03        &     27.06       &     13.43        &      8.19      &  2.28           &         36.52      &      22.24         &      38.23         &     18.17           \\ \Xhline{1.2pt}
\end{tabular}
\label{lab:Generalization_analysis}
\end{table}

\begin{table}[t]
\centering
\footnotesize
\caption{Comparison of different source pools. A: Market + Duke + MSMT.
B: A + UnrealPerson.
C: B + RandPerson + PersonX. All: C + CUHK03 + RAiD + VIPeR + PKU-Reid. SnP is consistently effective for different pools. 2\% IDs are selected from the source pool. Notations and evaluation metrics are the same as those in the previous table. }
\setlength{\tabcolsep}{1.3mm}
\begin{tabular}{l|cc|cc|cc|cc}
\Xhline{1.2pt}
\multirow{2}{*}{\begin{tabular}[c]{@{}l@{}}Search\\ method\end{tabular}} & \multicolumn{2}{c|}{Pool A} & \multicolumn{2}{c|}{Pool B} & \multicolumn{2}{c|}{Pool C} & \multicolumn{2}{c}{All} \\ \cline{2-9} 
                                                                         & R1           & mAP          & R1           & mAP          & R1           & mAP          & R1         & mAP        \\ \hline
Random                                                                   & 11.47        & 4.01       & 21.56        & 8.61        & 23.60        & 9.39        & 23.67      & 9.80       \\
Greedy                                                                   & 8.04        & 2.81        & 18.66        & 7.63        & 24.26        & 9.56        & 33.22      & 15.10      \\
SnP                                                                      & 14.17        & 4.47        & 25.44        & 10.16        & 36.32        & 17.22        & 38.23      & 18.17      \\ \Xhline{1.2pt}
\end{tabular}
\label{lab:different_source_pool}
\end{table}

\section{Further Analysis}
\label{sec:further_analysis}



\begin{figure}[t]
\begin{center}
	\includegraphics[width=1\linewidth]{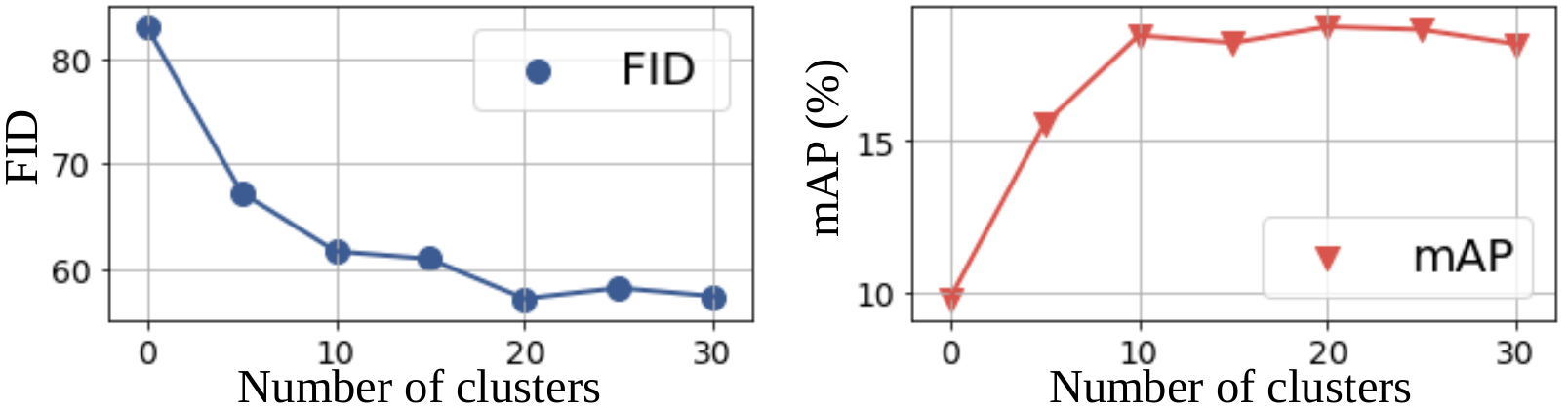}
\end{center}
\caption{Impact of the number of clusters $J$ to the domain gap (FID) between searched and target (left), and re-ID accuracy (right). AlicePerson is targeted, and the IDE~\cite{zheng2016mars} model is trained. }
\label{fig:cluster_num}
\vspace{-1em}
\end{figure}

\textbf{Generalization analysis of the training set generated by SnP.} In Table~\ref{lab:Generalization_analysis}, we directly apply the re-ID model obtained from the AlicePerson-specific training set to various test sets, such as Market, Duke, and MSMT. Compared with the model trained on the source pool, our target-specific model has higher accuracy on the target test set, and lower accuracy on other domains (MSMT for example shown in Table~\ref{lab:Generalization_analysis}). It demonstrates the generated training set is target-specific, improving re-ID accuracy on the given domain, while its generalization ability is weakened. But still, its application is promising given its target-specificity strength. 

\textbf{Analysis of different source pools.} We compare various source pools when AlicerPerson is the target domain. From Table~\ref{lab:different_source_pool}, we observe that our method gives very stable accuracy when using different source pools. In comparison, the compared methods, \textit{i.e.}, random sampling and greedy sampling, as much less stable. For example, random sampling performs well under Pool A which only contains real-world images and might be similar to the target domain. But it is much poorer under Pool C which has more synthetic data, meaning a larger domain gap from the target. 

\textbf{Analysis of hyperparameter $J$.} In Fig.~\ref{fig:cluster_num}, we show the increase of the number of clusters $J$ lowers the domain gap between the target and searched training set and generally increases the task accuracy. It reaches saturation after $J=20$. In this paper, we use $J=50$ in all setups. 

\textbf{Correlation between the number of images and training set quality.} While the correlation between the number of IDs and training set quality has been shown in Fig.~\ref{fig:correlation_study}. We show that when the number of ID is fixed, the correlation between the number of images and the rank-1 accuracy in Fig.~\ref{fig:fps_vs_random}. when we have a fixed number of IDs, the increase in the number of images brings a significant increase in rank-1 accuracy.

\begin{figure*}[t]
\centering
\includegraphics[width=\linewidth]{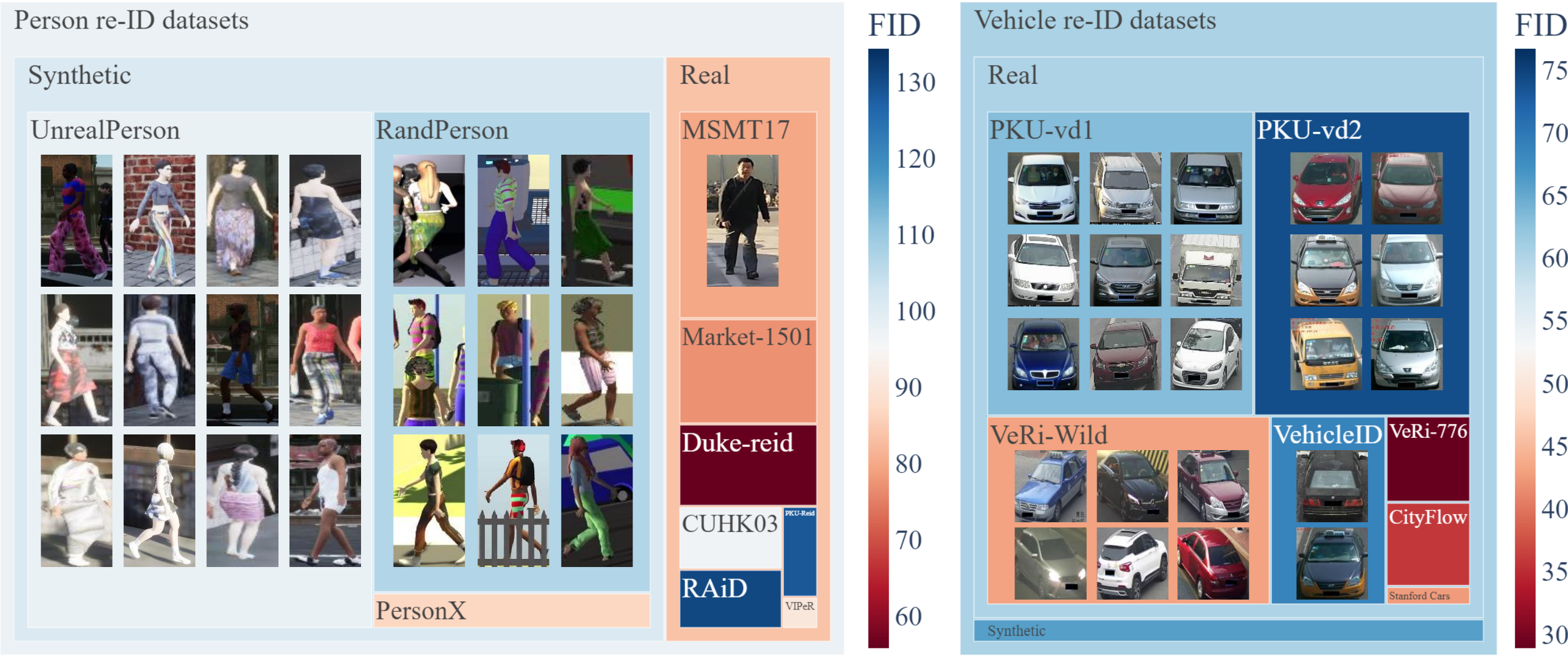}
\caption{The composition (tree map) of the source pool for the training set search. (\textbf{Left}:) Person re-ID datasets were collected for constructing the source pool. (\textbf{Right}:) Vehicle re-ID datasets. For both person re-ID and vehicle re-ID, we use synthetic data and real data. Each innermost rectangle represents a publicly available object re-ID dataset. The relative size of rectangles denotes the relative image size of datasets. The relative color shows the domain gap (measured in FID) value to the target test set. In person re-ID, the target for calculating the domain gap is AlicePerson~\cite{alice_benchmark}. In vehicle re-ID, the target for calculating the domain gap is AliceVehicle~\cite{alice_benchmark}.     }
\label{fig:meta_dataset}
\vspace{-0.5em}
\end{figure*}

\section{Source Pool Details}
\label{subsec:source_pool}

We show the composition (tree map) of the source pool for the training set search in Fig.~\ref{fig:meta_dataset}. On the left of Fig.~\ref{fig:meta_dataset}, for person re-ID, we summarize the details of each dataset in rectangles and show some examples. The size of each rectangle refers to the number of images in the dataset, while the color of the rectangle indicates the domain gap of that dataset to the target set. 
Regarding the target set in person re-ID, we use AlicePerson~\cite{alice_benchmark} for calculating the FID. 

We summarize the details of vehicle re-ID datasets in the right of Fig.~\ref{fig:meta_dataset}. The size and color of each rectangle carry the same meanings as those on the left of Fig.~\ref{fig:meta_dataset}. 
The AliceVehicle~\cite{alice_benchmark} is used as the target set for calculating FID. 

The datasets we use for the source pool are standard benchmarks, which are publicly available. We list their open-source as follows. For person re-ID:

\noindent \textbf{Market}~\cite{zheng2015scalable} \url{https://zheng-lab.cecs.anu.edu.au/Project/project_reid.html};

\noindent \textbf{Duke}~\cite{zheng2017unlabeled} \url{https://exposing.ai/duke_mtmc/};

\noindent \textbf{MSMT17}~\cite{wei2018person} \url{http://www.pkuvmc.com/publications/msmt17.html};

\noindent \textbf{CUHK03}~\cite{li2014deepreid} \url{https://www.ee.cuhk.edu.hk/~xgwang/CUHK_identification.html};

\noindent \textbf{RAiD}~\cite{das2014consistent} \url{https://cs-people.bu.edu/dasabir/raid.php};

\noindent \textbf{PersonX}~\cite{sun2019dissecting} \url{https://github.com/sxzrt/Instructions-of-the-PersonX-dataset};

\noindent \textbf{UnrealPerson}~\cite{zhang2021unrealperson} \url{https://github.com/FlyHighest/UnrealPerson};

\noindent \textbf{RandPerson}~\cite{wang2020surpassing} \url{https://github.com/VideoObjectSearch/RandPerson};

\noindent \textbf{PKU-Reid}~\cite{ma2016orientation} \url{https://github.com/charliememory/PKU-Reid-Dataset};

\noindent \textbf{VIPeR}~\cite{cho2018pamm} \url{https://vision.soe.ucsc.edu/node/178}.

\noindent For vehicle re-ID, we also list their open-source:

\noindent \textbf{VeRi}~\cite{liu2016large} \url{https://github.com/JDAI-CV/VeRidataset};

\noindent \textbf{CityFlow}~\cite{tang2019cityflow} \url{https://www.aicitychallenge.org/};

\noindent \textbf{VehicleID}~\cite{liu2016deep} \url{https://www.pkuml.org/resources/pku-vehicleid.html};

\noindent \textbf{VeRi-Wild}~\cite{lou2019veri} \url{https://github.com/PKU-IMRE/VERI-Wild};

\noindent \textbf{VehicleX}~\cite{yao2019simulating} \url{https://github.com/yorkeyao/VehicleX};

\noindent \textbf{Stanford Cars}~\cite{krause20133d} \url{http://ai.stanford.edu/~jkrause/cars/car_dataset.html};

\noindent \textbf{PKU-vd1}~\cite{yan2017exploiting} \url{https://www.pkuml.org/resources/pku-vds.html};

\noindent \textbf{PKU-vd2}~\cite{yan2017exploiting} \url{https://www.pkuml.org/resources/pku-vds.html}.

\clearpage
{\small
\bibliographystyle{ieee_fullname}
\bibliography{Content}
}

\end{document}